\documentclass[journal]{IEEEtran}
\usepackage{cite}
\usepackage{amsmath,amssymb,amsfonts}
\usepackage{algorithmic}
\usepackage{graphicx}
\usepackage{textcomp}
\usepackage{threeparttable}
\usepackage{color} 

\begin{document}

\title{Visual Anomaly Detection via Dual-Attention Transformer and Discriminative Flow}

\author{Haiming Yao,~\IEEEmembership{Student Member,~IEEE,}
        Weiluo,~\IEEEmembership{Student Member,~IEEE,}
        Wenyong Yu,~\IEEEmembership{Member,~IEEE,}
        
\thanks{Manuscript received XX XX, 20XX; revised XX XX, 20XX. This study was supported in part by the National Natural Science Foundation of China (Grant No. 51775214) (Corresponding author: Wenyong Yu.)}
\thanks{Haiming Yao is with the State Key Laboratory of Precision Measurement Technology and Instruments, Department of Precision Instrument, Tsinghua University, Beijing 100084, China. (e-mails: yhm22@mails.tsinghua.edu.cn).}
\thanks{Wei Luo and Wenyong Yu are with the State Key Laboratory of Digital Manufacturing Equipment and Technology, School of Mechanical Science and Engineering, Huazhong University of Science and Technology, Wuhan 430074, China(e-mails:ywy@hust.edu.cn; u201910709@hust.edu.cn).}}

\markboth{Submission to IEEE TRANSACTIONS ON INDUSTRIAL INFORMATICS}%
{Shell \MakeLowercase{\textit{et al.}}: A Sample Article Using IEEEtran.cls for IEEE Journals}

\maketitle
\begin{abstract}
In this paper, we introduce the novel state-of-the-art Dual-attention Transformer and Discriminative Flow (DADF) framework for visual anomaly detection. Based on only normal knowledge, visual anomaly detection has wide applications in industrial scenarios and has attracted significant attention. However, most existing methods fail to meet the requirements. In contrast, the proposed DADF presents a new paradigm: it firstly leverages a pre-trained network to acquire multi-scale prior embeddings, followed by the development of a vision Transformer with dual attention mechanisms, namely self-attention and memorial-attention, to achieve two-level reconstruction for prior embeddings with the sequential and normality association. Additionally, we propose using normalizing flow to establish discriminative likelihood for the joint distribution of prior and reconstructions at each scale. The DADF achieves 98.3/98.4 of image/pixel AUROC on Mvtec AD; 83.7 of image AUROC and 67.4 of pixel sPRO on Mvtec LOCO AD benchmarks, demonstrating the effectiveness of our proposed approach.
\end{abstract}


\section{Introduction}
\label{sec:introduction}

\IEEEPARstart{v}{isual} anomaly detection (VAD) has become increasingly popular in the industrial informatics community due to its non-destructive and precise detection capabilities. It plays a vital role in achieving quality control and fault diagnosis for various products\cite{r1}, which is critical for intelligent manufacturing.

Initially, researchers attempted to identify various anomalies using supervised methods\cite{r2}. However, the arduous task of collecting and labeling anomaly samples has prompted researchers to shift their focus to unsupervised approaches. Under such training fashion, individuals only need to collect normal samples and establish a normal distribution. Subsequently, they can utilize the discrepancies between the samples and the normal distribution to evaluate the probability of anomalies. Particularly after the release of Mvtec AD benchmark\cite{r3}, significant progress has been made with these types of methods.

Overall, current unsupervised VAD methods have formed different branches with certain structural characteristics. These branches include reconstruction-based methods\cite{r4}, embedding-based methods\cite{r5}, regression-based methods\cite{r6}, and likelihood-based methods\cite{r7}. Reconstruction-based methods are trained to reconstruct the normal samples during the training stage and expect to reveal anomalies by exhibiting large reconstruction error for anomalous samples. Regression-based methods, on the other hand, attempt to identify anomalies by regressing on the regression errors of anomalous samples. Furthermore, embedding-based methods aim to achieve detection by computing the distance between the embedded features of anomalous and normal samples, while likelihood-based methods model normalcy through probability.

\begin{figure}[t]\centering
\includegraphics[width=8.8cm]{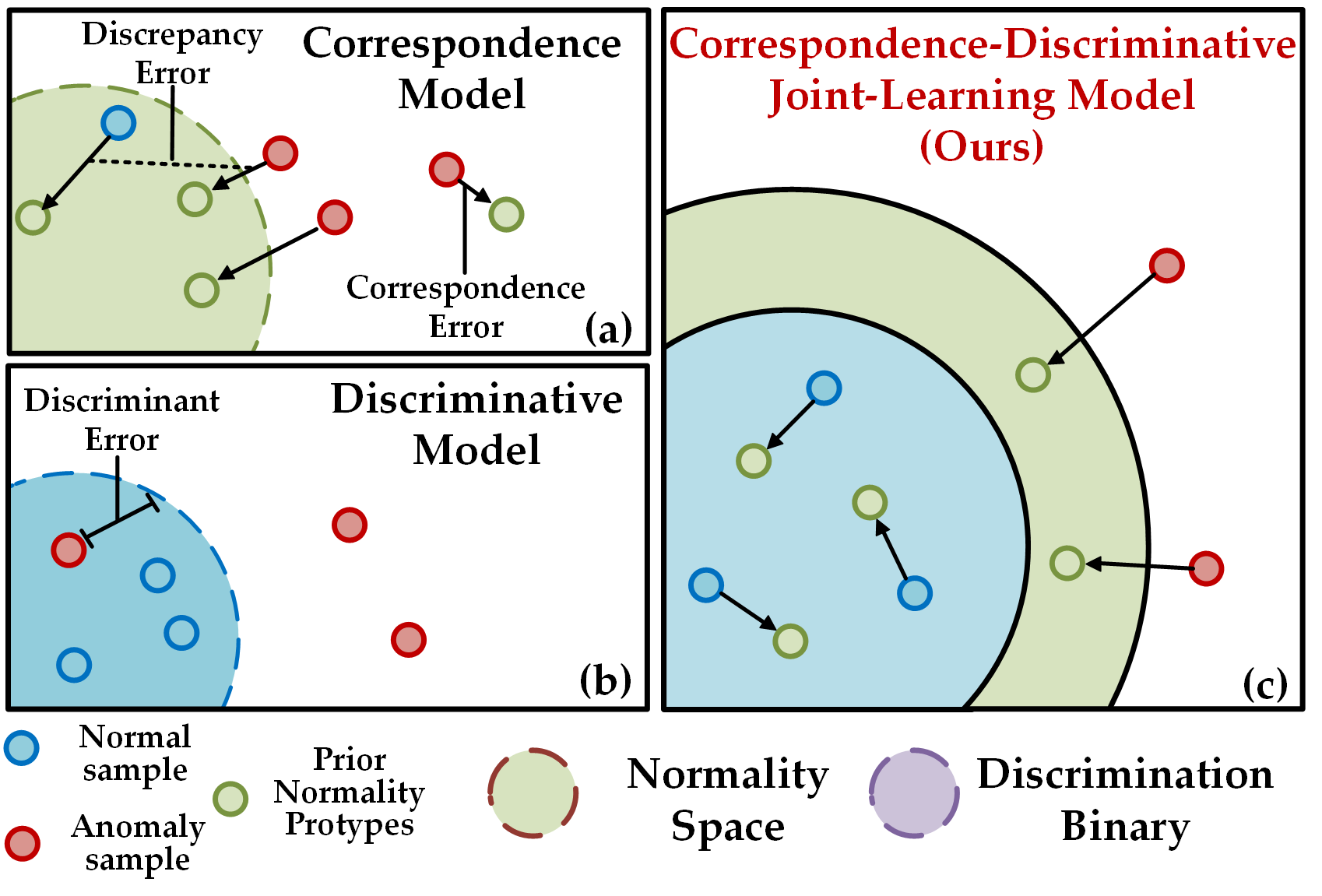}
\caption{Schematic diagram of two intrinsic working mechanisms. (a) Correspondence errors and difference errors in the correspondence mechanism can lead to misjudgments. (b) In the discriminant mechanism, misjudgment may occur due to insufficient discriminative properties of features. (c) The proposed mechanism of the correspondence-discriminant joint effect. It first learns a robust correspondence mechanism and then uses the correspondence relationship to enable a more discriminative capacity of model.}
\label{FIG1}
\end{figure}

To delve deeper, the aforementioned methods can be further traced back to the working mechanisms of correspondence and discrimination as depicted in Fig. 1. The correspondence mechanism, which is utilized in reconstruction-based and regression-based methods, involves finding prior normal prototypes as references and using their discrepancy as the judgment criterion. However, this approach is not always reliable as over-generalization or over-reconstruction can result in prototypes belonging to the anomalous space, leading to correspondence failure. Even when the prototypes are appropriate, correspondence discrepancy errors can still result in misjudgment.  In contrast, the discrimination mechanism involves the dense clustering of normal features in a deep subspace, as seen in embedding-based models' memory bank and likelihood models' density distribution. This approach requires high discriminative capacity in feature representations, as insufficient discriminability between normal and anomalous features can lead to discrimination failure.

Drawing from the preceding analysis, we put forth a novel approach to enhance the model's performance by capitalizing on the synergistic interplay between the intrinsic mechanisms outlined above. Specifically, our proposal involves the model learning a robust correspondence mechanism followed by employing the correspondence relationship for discrimination modeling. By transforming the primary criterion of the corresponding discrepancy into discrimination, we can enhance the judgment accuracy of the corresponding mechanism. Likewise, by utilizing the corresponding relationship as a discriminative criterion in addition to the representative nature of prior features, we can significantly augment the discriminative capacity of the model.

To validate our proposal, this study introduces a novel architecture called Dual Attention Vision Transformer and Discriminative Flow (DADF). In line with existing methods, we first extract features from a pre-trained network to create a multi-scale prior embedding. Next, we introduce a novel vision Transformer that utilizes a dual-branch approach, integrating self-attention and memorial-attention mechanisms to establish a reliable correspondence of a two-level reconstruction with sequential and normality associations. Instead of using the two reconstruction discrepancies as anomaly indicators, we present a novel approach that utilizes normalizing flow to learn a discriminative model over the joint distribution of the original features and two reconstructed features, resulting in the normality likelihood. The collaborative effect of the correspondence-discrimination mechanism can improve performance significantly. Our core contributions can be summarized into three aspects.

\begin{enumerate}
\item This research proposes a novel approach to enhance VAD by utilizing the collaborative effect of correspondence and discriminative mechanisms. Based on this motivation, we introduce a new framework, the Dual-Attention vision Transformer and Discriminative Flow (DADF), which can produce advanced anomaly detection performance.

\item Our method features two key designs: firstly, a Dual-Attention visual Transformer is proposed incorporating self-attention and memorial-attention mechanisms to establish stable correspondence of a two-level reconstruction with sequential and normality associations. Secondly, a discriminative normalizing flow is developed to learn the normality likelihood over the joint distribution of prior and corresponding reconstructed features.

\item Our method is assessed on various industrial benchmarks, and the results show that it achieves a state-of-the-art anomaly segmentation performance of 98.4 AUCROC and an anomaly detection performance of 98.3 AUCROC on the Mvtec AD dataset. Additionally, it achieves an anomaly segmentation performance of 67.4 sPRO and an anomaly detection performance of 83.7 AUCROC on the MvtecLOCO AD dataset.
\end{enumerate}

The remainder of this paper is structured as follows: In Section II, we review recent works related to anomaly detection. In Section III, we elaborate our DADF method in detail. In Section IV, we present comprehensive experiments, and in the final section, we conclude the paper and provide future perspectives.

\section{Related work}

In this section, we will present and discuss recent methods for VAD that operate on different working mechanisms, including reconstruction-based and regression-based methods that utilize correspondence mechanisms, as well as embedding-based and likelihood-based methods that rely on discriminative mechanisms.

\subsection{Correspondence mechanisms enabled methods}

Reconstruction-based methods using autoencoder (AE) models have been a popular and active area of research in the VAD community. This baseline model has been improved with various approaches, including memory-augmented AE\cite{r4}\cite{r8} that uses a memory bank to construct normal prototypes for better suppression of abnormal reconstructions. Other optimization techniques such as perceptual loss\cite{r9}, structural similarity AE\cite{r10}, and integration with generative adversarial networks have also been applied. Regression-based models using knowledge distillation\cite{r6} have also been proposed, relying on regression error for anomaly detection. Multi-resolution knowledge distillation\cite{r11} and reverse distillation\cite{r12} have been proposed to improve this approach. Simulated defects\cite{r13} have also been introduced to suppress over-generalization to anomalies.

\subsection{Discriminative mechanisms based methods}

Discriminative mechanisms-based methods have been shown to be effective in VAD by leveraging the semantic extraction ability of deep neural networks. Embedding-based models use the distance between samples and the normal cluster in the latent space to detect anomalies. PADIM\cite{r14} fits a multivariate Gaussian distribution and utilizes the Mahalanobis distance for normality detection. SPADE\cite{r15} stores normal features and calculates the distance of retrieved nearest features as the anomaly score during the testing phase. This approach is further improved in Patchcore\cite{r5}. Recently, normalization flow (NF)\cite{r16} using reversible neural networks has been introduced in VAD\cite{r17}\cite{r7}. NF has bijective mappings and tractable Jacobian determinants, which enable the establishment of a connection between the prior distribution of normal features and the standard normal distribution. Abnormal features are projected outside of the learned distribution.


\begin{figure*}[t]
\centerline{\includegraphics[width=\textwidth]{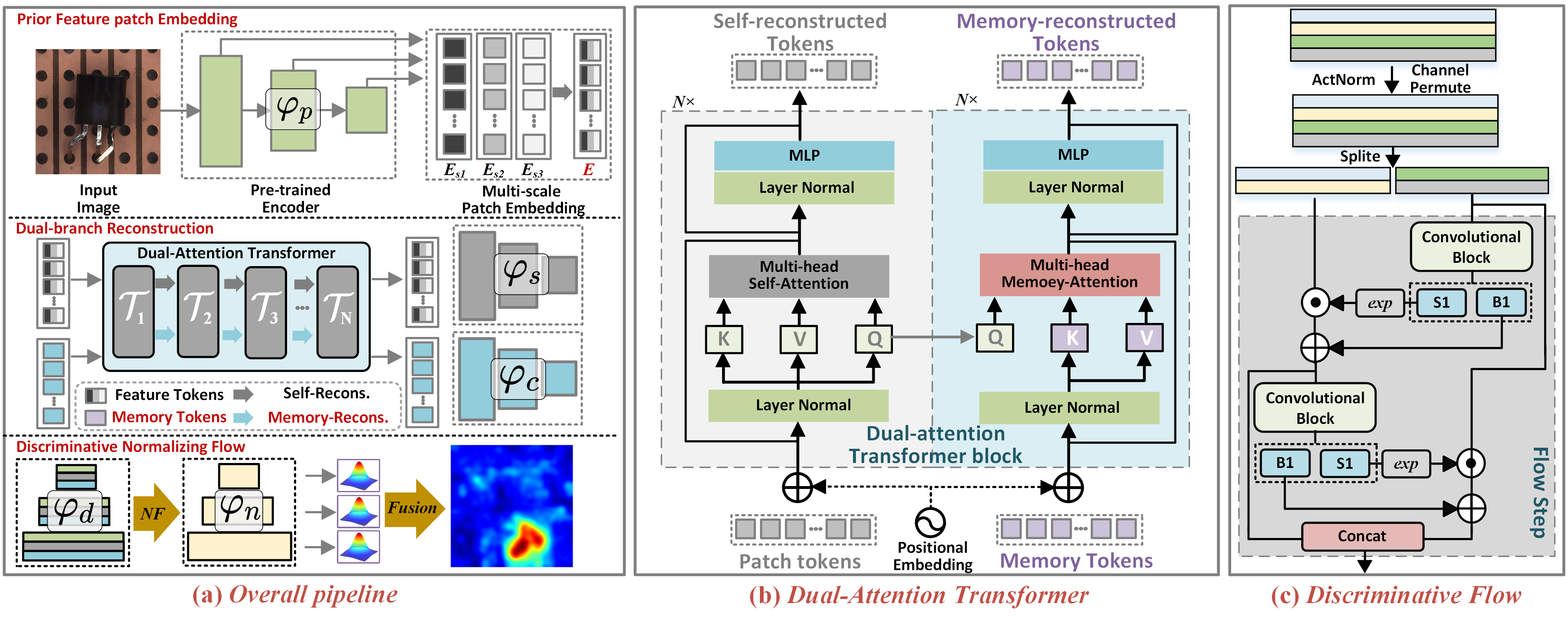}}
\caption[width=\textwidth]{
(a) The overall pipeline of the proposed DADF comprises three main sub-steps. Firstly, multi-scale prior feature embeddings are extracted from the pre-trained encoder network on the input image. Subsequently, the dual-attention transformer performs two-level reconstructions to obtain self-reconstruction and memory reconstruction features. Lastly, the proposed discriminative flow is utilized to estimate normality on the joint distribution of prior and reconstructed features, resulting in robust anomaly localization. (b)
The dual-attention transformer structure. Apart from the self-attention reconstruction pathway, it additionally incorporates memory tokens for reconstruction based on memory attention. (c). The affine coupling structure of the invertible transformations blocks for normalizing flows\cite{r16}

}
\label{fig1}
\end{figure*}
\section{Proposed DADF methodology}

\subsection{Overall Schema}

The DADF proposed in this study is presented in Fig. 2(a) and involves three main steps. Firstly, multi-scale prior feature embeddings are extracted from pre-trained CNNs at different scales. Secondly, a vision Transformer model with self-memorial dual attention branches is employed for two-level reconstruction of the prior embeddings using self and memory-based attention mechanisms, as shown in Fig. 2(b). Finally, discriminative normalizing flows are incorporated at each scale to learn the normality likelihood of the joint distribution of the two-level reconstructed features and prior features, as illustrated in Fig. 2(c). During testing, anomalous samples display significant reconstruction errors in the two-level reconstructions produced by the dual-attention visual Transformer. As a result, the joint distribution of the anomalous samples deviates from the learned distribution during training, leading to a lower normality likelihood, which serves as a robust criterion for anomaly detection.
 
\subsection{Dual Attention vision Transformer}

\subsubsection{Multi-scale Prior Feature Embedding} 

As presented in Fig.2(a), given an input image, we first fed it into a CNN model pre-trained on the ImageNet dataset to extract multi-scale prior feature embeddings. Specifically, the prior features at different levels of convolutional modules are extracted. Then, with different projection heads, we use different patch sizes $P_{i}$ to embed prior features $\boldsymbol{\phi_{\mathcal{P}}}=\big\{ \phi_{\mathcal{P}}^{(i)}\in \mathbb{R} ^{H_{i}\times W_{i}\times C_{i}},i=1,...,N  \big\}$ from different levels into 1D sequences of token embeddings $\left \{  E_{1},E_{2},...,E_{N} \mid E_{i}\in \mathbb{R}^{L\times D} \right \}$ with the dimension $D$ and same length $L$. Finally, we concatenate tokens from different scales in the embedding dimension to obtain multi-scale prior feature embeddings $\mathbf{E}\in \mathbb{R}^{L\times ND}$.

\subsubsection{Self-Memorial Two Branch Attention}

In this research, we modify the vision Transformer to reconstruct prior features by exploiting its intrinsic capability to model global representations and capture long-range relations. Nevertheless, we have identified two constraints when using the vanilla Transformer for this purpose. First, residual connection paths may cause anomalous information leakage. Second, due to the tendency of anomalies to be linked to their own or neighboring tokens with similar anomalous patterns, it is difficult to establish global connections between anomalies and the entire sequence. Both of these inductive biases can result in unfavorable shortcut identity mapping for reconstruction-based VAD.

Going beyond previous methods, we propose a novel Transformer structure with two attention branches, as illustrated in Fig. 2(b). We introduce an additional set of learnable tokens with the same shape and position encoding as the feature tokens. Then two sets of tokens are propagated in the dual attention Transformer. Finally, the self/memory two-level reconstruction results can be obtained.

To conduct a more comprehensive examination, as shown in Fig. 3, the model's self-attention branch is designed to capture sequential associations from the raw feature token sequence, which is expected to be most effective for the reconstruction task. It's worth noting that for the memorial-attention branch, the query is obtained from the original patch token sequence, while the key-value pairs are derived from the memory token sequence. The memory tokens are supervised during the training phase to reconstruct normal feature tokens via memorial associations, which is comparable to using feature tokens as queries to retrieve the most fitting key-value pair bank generated by memory tokens for the training normal pattern. In this process, the memory branch is tasked with implicitly producing a key-value pair bank with normal patterns that can appropriately correspond to the query. Consequently, during the testing phase, the query tokens of abnormal features only allow access to the learned normal patterns, thus impeding the reconstruction of anomalies.

Similarly, we utilize a series of projection heads placed after the Transformer model to align the dimensions of the reconstructed two sets of tokens, denoted as $\boldsymbol{\phi_{\mathcal{S}}}$ and $\boldsymbol{\phi_{\mathcal{M}}}$, with those of the prior features, as denoted in Fig.2 (a).
\begin{figure}[t]
\centerline{\includegraphics[width=8.8cm]{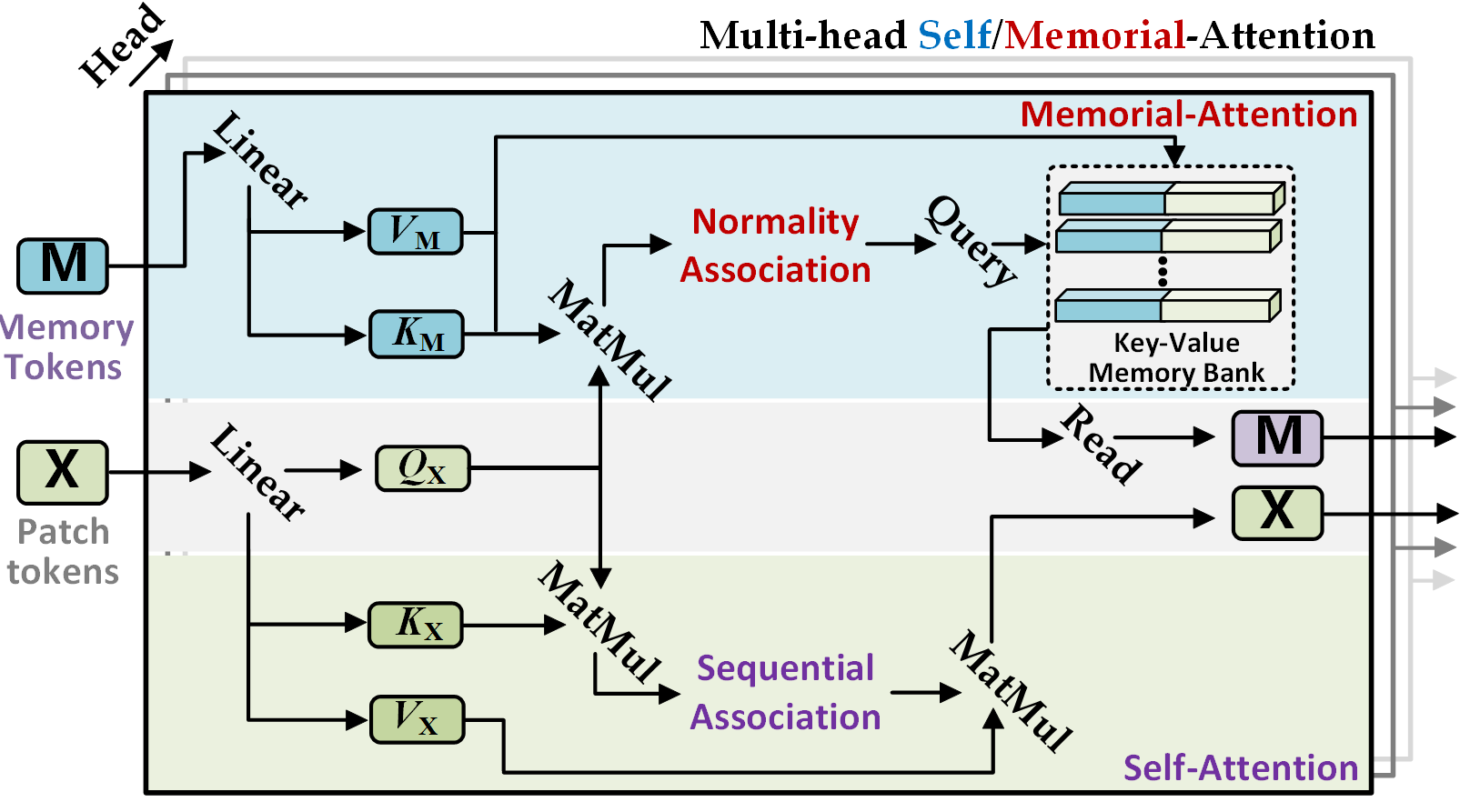}}
\caption[width=8.8cm]{Dual-attention architecture. Dual-Attention consists of memorial- and self-attention, which models the normality association and sequential association simultaneously.
}
\label{fig1}
\end{figure}

\subsection{Discriminative normalizing flow}

After obtaining the two-level reconstruction outcomes, a straightforward reconstruction discrepancy could have been used as an anomaly criterion. However, as previously mentioned, this approach may limit detection performance. To address this, a novel discriminative normalization flow for normality modeling is introduced in this study.

With the NF model $g\left ( \boldsymbol{\theta} \right ) $ possessing the capacity of a bijective invertible mapping, we are able to establish an invertible mapping between any arbitrary given distribution $p_{U}\left ( \boldsymbol{u} \right ) $ and the base distribution $p_{Z}\left ( \boldsymbol{z} \right )\sim \mathcal{N} \left (  \boldsymbol{0}, \boldsymbol{I} \right )$ as:
\begin{equation}
\mathrm{log} p_{Z}\left ( \boldsymbol{z} \right ) = \mathrm{log} p_{U}\left ( \boldsymbol{u} \right )+\mathrm{log}\left | \mathrm{det} \bigtriangledown_{\boldsymbol{z}}g( \boldsymbol{z}, \boldsymbol{\theta}) \right | 
\end{equation}

Therefore, as illustrated in Fig. 4(a), we can utilize the prior distribution of normal features to optimize the NF model for establishing the likelihood of normality. Consequently, anomalies that deviate from the normal prior distribution will possess a relatively lower normality likelihood. Nevertheless, prior research has indicated that the NF is predisposed to fixate on local information, prompting existing methodologies to employ pre-trained features to gather semantics. However, this approach imposes significant demands on the normal/abnormal discriminant of the feature extraction network. The discriminative capacity of the feature extraction network consequently exerts a direct influence on the assessment of normality likelihood.

Hence, we introduce the novel discriminative NF to enhance feature discriminability by using reconstruction discrepancies. As illustrated in Fig. 4(b), instead of utilizing solely the prior features $\boldsymbol{\phi_{\mathcal{P}}}$, we leverage their concatenated effect with the two-level reconstruction results $ \boldsymbol{\phi_{\mathcal{D}}}= \left \{\boldsymbol{\phi_{\mathcal{P}}}, \boldsymbol{\phi_{\mathcal{S}}},\boldsymbol{\phi_{\mathcal{M}}}  \right \} $. 
By utilizing the significant differences in the reconstruction behavior between the normal and abnormal features, we can achieve a more reliable and effective discriminative criterion than relying solely on the prior features. This can, in turn, improve the difference in normality likelihood between normal and anomalous samples. 

\begin{figure}[t]
\centerline{\includegraphics[width=8.8cm]{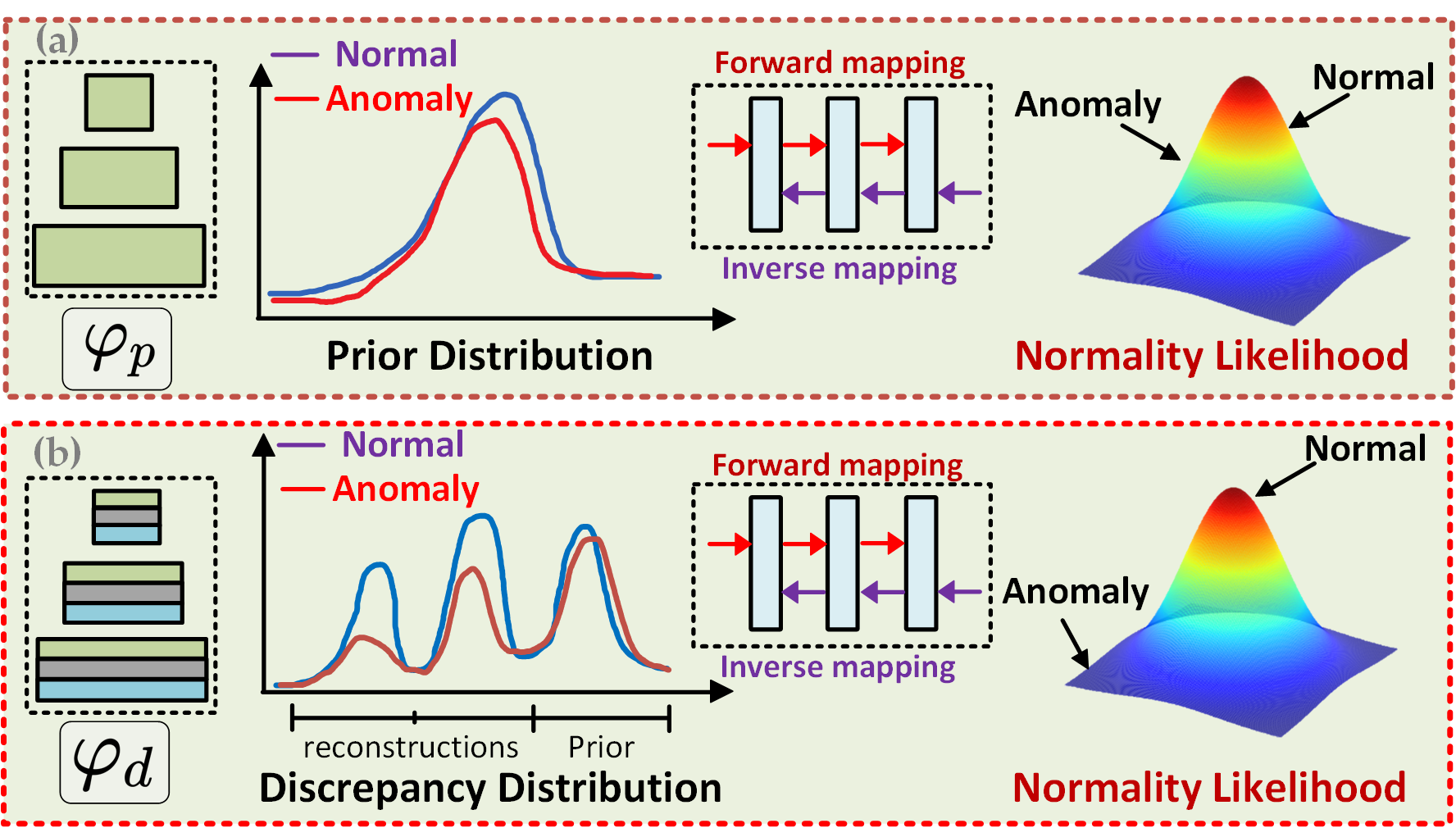}}
\caption[width=8.8cm]{
Comparison of vanilla normalizing flow and the proposed discriminative flow. Our approach provides more reliable normality estimates. }
\label{fig1}
\end{figure}

Thus, the log-likelihood of the distribution of the proposed discriminative flow can be expressed as:
\begin{equation}
\mathrm{log} \hat{p}_{\phi_{\mathcal{D} }} \left ( \boldsymbol{\phi_{\mathcal{D}}}, \boldsymbol{\theta _{\mathcal{D}}} \right )  = \sum_{i=0}^{N}\left \{ \mathrm{log}p_{Z}\left ( \boldsymbol{z} \right )+ \mathrm{log}\left | \mathrm{det}\boldsymbol{J}^{(i)}  \right | \right \}
\end{equation}
where the $\boldsymbol{J}^{i} = \bigtriangledown _{\phi_{d}^{(i)}}g^{-1}_{(i)}\big (  \boldsymbol{\phi_{\mathcal{D}}}^{(i)}, \boldsymbol{\theta_{\mathcal{D}}^{(i)} }\big ) $ is the Jacobian matrix of the 
$i$-th bijective invertible NF model parameterized by $\boldsymbol{\theta}_{\mathcal{D}}^{(i)}$.

\subsection{Training Objective}
The main focus of the DADF model in this study encompasses two primary objectives, namely, prior reconstruction and discriminative normality likelihood estimation. 

To achieve the reconstruction objective, we employ the squared Frobenius norm as the optimization objective:
\begin{equation}
\mathcal{L}_{\mathcal{S}}\left (  \boldsymbol{\theta }_{\mathcal{T} } \right ) = \sum_{i=1}^{N}\Big\{   \sum_{h=1}^{H_{i}}  \sum_{w=1}^{W_{i}} \left \|  {\large\varphi_{\mathcal{P}}^{(i)}} \left ( h,w \right) -{\large\varphi_{\mathcal{S}}^{(i)}} \left ( h,w \right)  \right \|_{2}^{2} \Big\}
\end{equation}
where the $\mathcal{L}_{\mathcal{S}}\left (  \boldsymbol{\theta }_{\mathcal{T} } \right )$ represents the self-reconstruction loss for dual-attention Transormer with parameter $ \boldsymbol{\theta }_{\mathcal{T}}$. Similarly, the memory-based reconstruction loss $ \mathcal{L}_{\mathcal{M}}$ is expressed as:
\begin{equation}
\mathcal{L}_{\mathcal{M}}\left (  \boldsymbol{\theta }_{\mathcal{T} } \right ) = \sum_{i=1}^{N}\Big\{   \sum_{h=1}^{H_{i}}  \sum_{w=1}^{W_{i}} \left \|  {\large\varphi_{\mathcal{P}}^{(i)}} \left ( h,w \right) -{\large\varphi_{\mathcal{M}}^{(i)}} \left ( h,w \right)  \right \|_{2}^{2} \Big\}
\end{equation}

For discriminative normal likelihood estimation, we optimize the NF model by a maximum likelihood objective, where the Kullback-Leibler (KL) divergence  $D_{KL}\big [ \hat{p}_{\phi_{\mathcal{D} }} \left ( \boldsymbol{\phi_{\mathcal{D} }}, \boldsymbol{\theta _{\mathcal{D}}}\right ) \left |  \right |  p^{*}_{\phi_{\mathcal{D} }} \left ( \boldsymbol{\phi_{\mathcal{D} }}\right )  \big ]$ is employed:
\begin{equation}
\mathcal{L}\left (  \boldsymbol{\theta }_{\mathcal{D} } \right ) =\sum_{i=0}^{N} \left ( \mathbb{E} \left [ { \hat{p}_{\phi_{\mathcal{D} }}^{(i)} \left ( \boldsymbol{\phi_{\mathcal{D}}^{(i)}}, \boldsymbol{\theta}_{\mathcal{D}}^{(i)} \right )}-  p^{*}_{\phi_{\mathcal{D} }^{(i)}} \left ( \boldsymbol{\phi_{\mathcal{D} }^{(i)}}\right ) \right ]  \right ) 
\end{equation}

To sum up, the comprehensive training loss of the model can be formulated as a combination of the aforementioned reconstruction loss and likelihood loss:
\begin{equation}
\mathcal{L}\left ( \boldsymbol{\theta } \right ) =\underbrace{\mathcal{L}_{\mathcal{S}}\left (  \boldsymbol{\theta }_{\mathcal{T} } \right ) +\mathcal{L}_{\mathcal{M}}\left (  \boldsymbol{\theta }_{\mathcal{T} } \right )}_{Reconstruction} +\underbrace{\mathcal{L}\left (  \boldsymbol{\theta }_{\mathcal{D} } \right )}_{Likelihood} 
\end{equation}
\subsection{Anomaly Score Function}

\begin{table*}
\caption{{The quantitative results of different methods in MVTec AD at the image/pixel-level}}
\label{table}
\setlength{\tabcolsep}{3pt}
\begin{threeparttable}
\begin{tabular}{p{\textwidth}}
$\includegraphics[width=\textwidth]{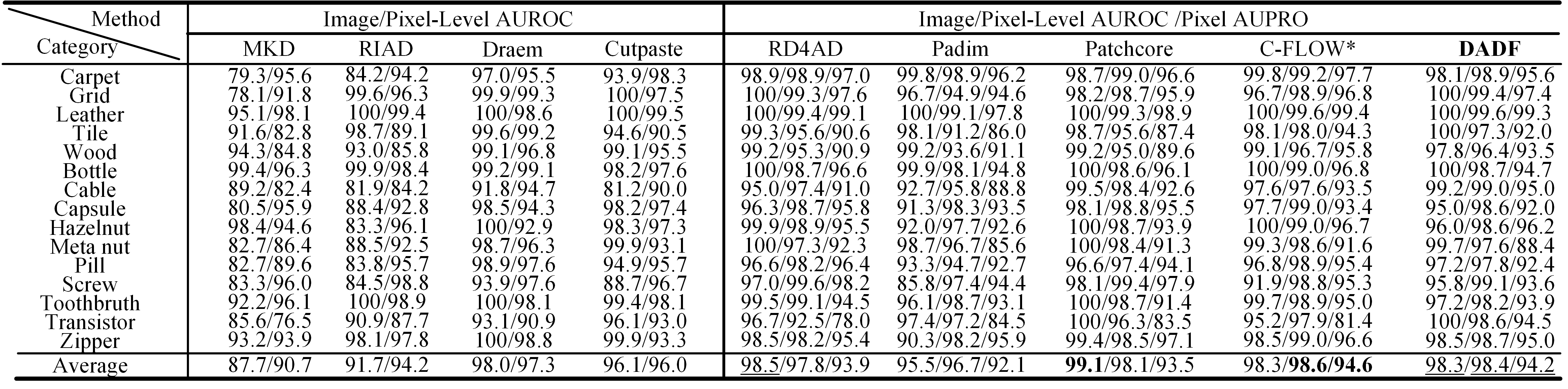}$
\end{tabular}
\begin{tablenotes}
       \footnotesize
       \item[1]The best performance is indicated by bold font, while the second best is indicated by an underline.
\end{tablenotes}
\end{threeparttable}
\label{table1}
\end{table*}

\begin{table*}
\caption{{The quantitative results of various methods on the Mvtec LOCO AD dataset. The results for each category are reported as logical anomalies/structural anomalies or the average of both. The overall averages are reported as logical anomalies/structural anomalies and the average of both. The comparison methods' results are obtained from sources \cite{r23,r24,r25}.}}

\label{table}
\setlength{\tabcolsep}{3pt}
\begin{threeparttable}
\begin{tabular}{p{\textwidth}}
$\includegraphics[width=\textwidth]{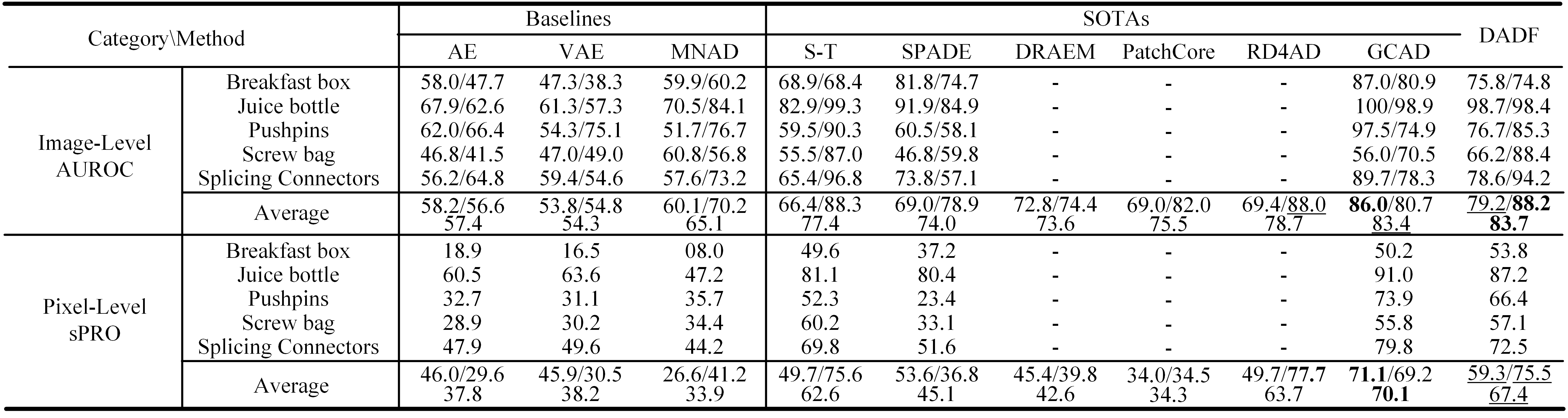}$
\end{tabular}
\end{threeparttable}
\label{table2}
\end{table*}

In the testing phase, the extracted prior embeddings are reconstructed using the dual-attention Transformer. Next, both the prior and reconstructed features are inputted into the discriminative normalizing flow in order to obtain their abnormal likelihood:
\begin{equation}
\mathcal{A}\left ( h,w \right )  = \sum_{i=0}^{N} \big (\mathbf{1}  - \left \|\Gamma \big( g^{-1}_{(i)}\big(  \boldsymbol{\phi_{\mathcal{D}}}^{(i)},\boldsymbol{\theta_{\mathcal{D}}^{(i)} }  \big) \big)\left ( h,w \right )  \right \|_{2}^{2} \big )  
\end{equation}
where the $\Gamma()$ dentoes the interpolation operation, and the max value of $\mathcal{A}$ can produce the image-level anomaly score.

\section{Experimental Results}
This section validates the DADF method via experiments, including benchmark comparisons and ablation studies for further performance analysis.

\subsection{Experiments Setup}

\subsubsection{Implementation Details}
Each image is resized to 256×256 resolution and normalized using mean and standard deviation from the ImageNet dataset. A pre-trained network on ImageNet is used as the prior feature extractor network, frozen during training. The default model is Wide-Resnet50, extracting multi-scale hierarchical feature maps from the outputs of the first three convolutional modules, with patch sizes for embedding are set to 4, 2, and 1, respectively. The DADF model is trained from scratch on only normal samples, using the AdamW optimizer with a learning rate of 1e-4 and a batch size of 8. The training strategy involves first training the dual-attention Transformer with the reconstruction loss for 300 epochs and then training the discriminative flow with the likelihood loss for 100 epochs. The structural configuration of the dual-attention Transformer follows the design of the MAE\cite{r18} encoder, with reduced embedding dimension to 480, depth to 8, and heads to 8 for efficient computation. For the discriminative flow, the affine-coupling structure in NICE is adapted as the NF models for each scale, employing a convolutional block consisting of a 3×3 depthwise separable convolution, a Leaky Relu activation, and a 1×1 convolution as two subnets. Experiments were performed on a computer with Xeon(R) Gold 6226R CPUs@2.90GHZ and two NVIDIA A100 GPUs with 40GB of memory.

\subsubsection{Dataset Descriptions}
Our DADF method was evaluated on two benchmarks, namely Mvtec AD and Mvtec LOCO AD \cite{r3,r19}. The former consists of 5 texture categories and 10 object categories with 3629 normal training images, and 467 normal and 1258 abnormal test images. The latter is a recently introduced dataset designed for detecting logical anomalies, and it contains a total of 1772 training samples, 304 validation samples, and 1568 testing samples with both structural and logical anomalies.

\subsubsection{Evaluation Metrics}

As per previous studies, we use the area under the receiver operating characteristic (AUROC) and normalized area under the per-region overlap curve (AU-PRO)\cite{r6} as the primary quantitative evaluation metrics. Additionally, the sPRO\cite{r23} metric is utilized for the Mvtec LOCO dataset. Higher values for all metrics indicate superior performance.

\subsection{Comparison with the State-of-The-Art models}
\subsubsection{Mvtec AD} We compared our method with several advanced methods on Mvtec AD: MKD\cite{r11}, RIAD\cite{r20}, Draem\cite{r21}, Cutpaste\cite{r22}, RD4AD\cite{r12}, Padim\cite{r14}, Patchcore\cite{r5}, and C-FLOW\cite{r7}. Table I presents the results of our comparative experiments, indicating the remarkable performance of the proposed DADF model. Our method achieves image/pixel-level AUROC scores of 98.3/98.4 across all 15 categories, ranking second in overall performance. Additionally, our approach achieves a Pixel-level AUPRO of 94.2, which is only outperformed by the C-FLOW method. However, it is worth noting that C-FLOW's performance was achieved with various backbones and resolutions, which is essentially an unfair comparison. In fact, when using the same backbone and resolution as our method, C-FLOW's pixel-level AUROC score reduces to 97.9, as stated in the original paper.

Overall, our method achieves state-of-the-art performance on Mvtec AD, with the exception of being at a disadvantage in comparison to C-FLOW due to an unfair comparison factor, and also having inferior image-level anomaly detection performance when compared to Patch Core.
\begin{figure}[t]
\centerline{\includegraphics[width=8.8cm]{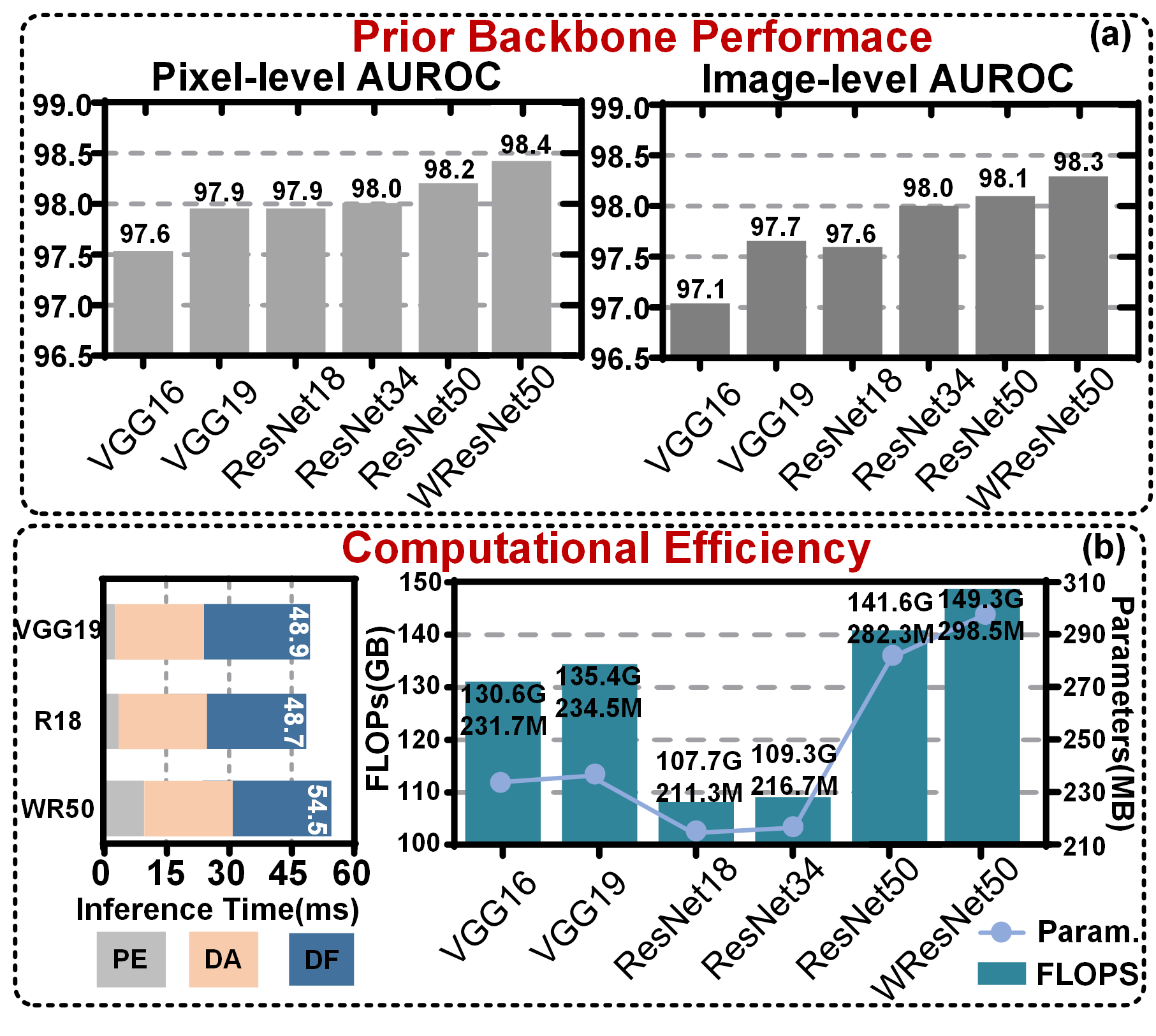}}
\caption[width=8.8cm]{
(a) Impact of prior feature extraction on model performance. (b) Computational efficiency analysis. "PE" refers to prior feature extraction, "DA" represents dual-attention transformer, and "DF" denotes discriminative flow.
}
\label{fig1}
\end{figure}

\subsubsection{Mvtec LOCO AD}

With the exception of GCAD\cite{r23}, a recently introduced method specifically tailored to Mvtec LOCO AD, there have been few new approaches to tackle this type of logical anomaly detection problem. Therefore, further exploration of this area is still warranted. We compared our method on itwith some existing methods, including baseline methods: f-AnoGAN\cite{r26}, AE\cite{r23}, VAE\cite{r27}, and MNAD\cite{r28}; state-of-the-art methods: S-T\cite{r6}, SPADE\cite{r15}, DREAM\cite{r21},  Pachcore\cite{r5}, RD4AD\cite{r12}, and GCAD\cite{r23}.

Table II displays the comparison results of various methods. Unlike Mvtec AD, the performance of all models drops remarkably on Mvtec LOCO AD. The baseline models achieve a maximum image-level AUROC of 65.1 and pixel-level sPRO of 38.2. Among the recent state-of-the-art methods, GCAD remains the best-performing method overall, followed by our DADF with a slight gap. Compared to other approaches, our DADF and GCAD demonstrate significant advantages, particularly for logical anomalies. For instance, our method demonstrates a +5.0 image-level AUROC and +3.7 pixel sPRO improvement compared to RD4AD, and a +8.2 image-level AUROC and +33.1 pixel sPRO improvement compared to Patchcore.

\begin{figure}[t]
\centerline{\includegraphics[width=8.8cm]{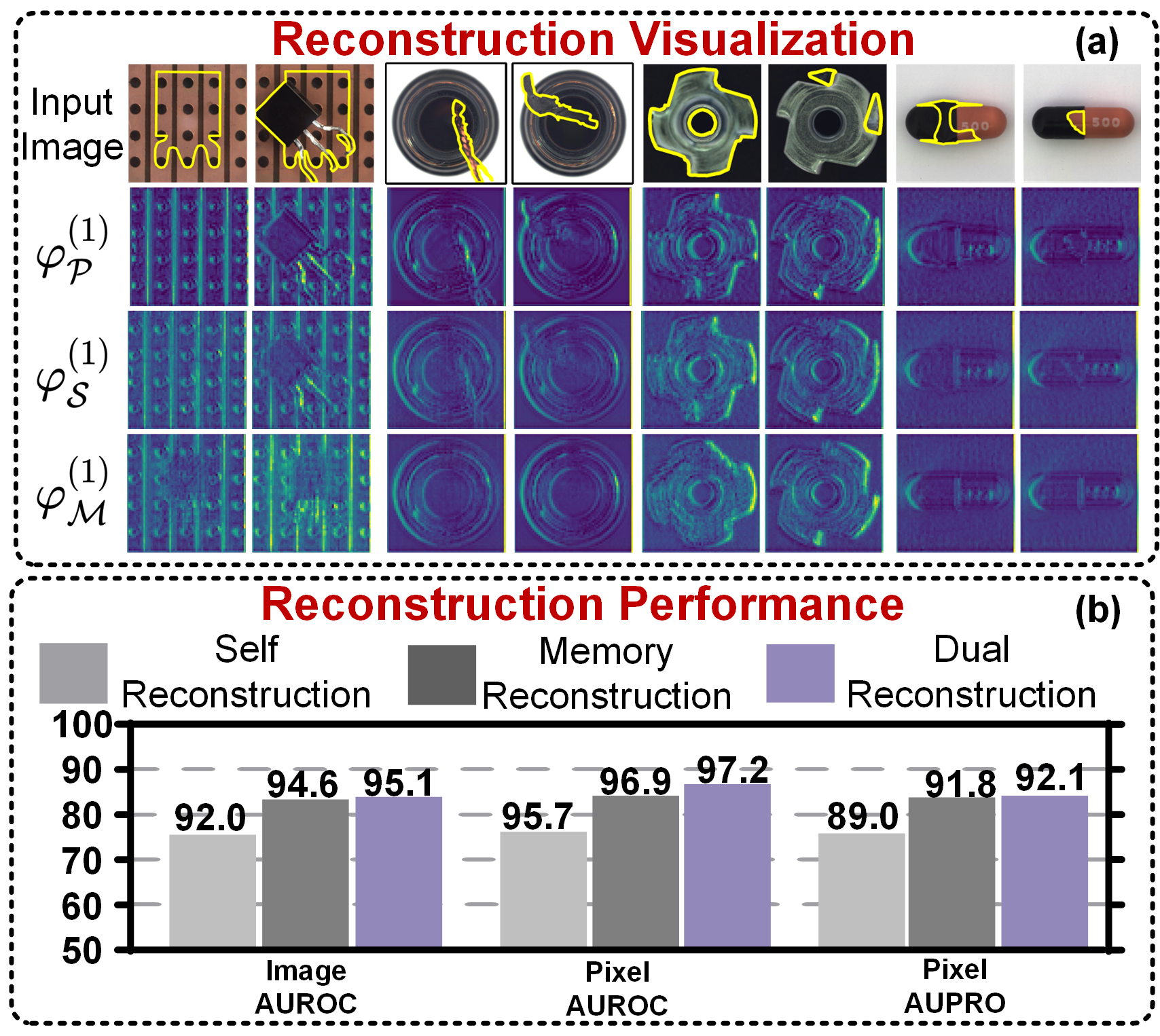}}
\caption[width=8.8cm]{
(a).Visualization results of prior feature and two reconstruction feature by the DADF. (b) Reconstruction-based Detection performance comparison of the different reconstruction approaches. 
}

\label{fig1}
\end{figure}

\subsection{Ablation studies}

\subsubsection{Influence of the prior feature extraction}

The discriminative abilities of feature extraction networks in detecting anomalies are a crucial factor that affects their performance. In our DADF approach, we leverage not only the discriminative properties of prior features but also the discrepancies in reconstruction results obtained from the dual-attention Transformer for improved discrimination. The experimental findings depicted in Fig. 5(a) demonstrate that the WResNet50 backbone network yields the best overall performance. Moreover, the approach achieves commendable detection accuracy of over 97.0 even for lightweight backbone networks such as VGG16, VGG19, and ResNet18.

On the right side of Fig. 5(b), we present the experimental results, which indicate that the computational complexity of our proposed DADF varies for different backbone networks. Our method incurs an efficient total computational cost for the lightweight ResNet18 backbone network, and an increase in backbone scale leads to a corresponding increase in the computational cost of the subsequent DA-DF processes. The inference time results on the left side of Fig. 5(b) reveal that our method is stable and exhibits an acceptable inference speed of approximately 50ms. Overall, the proposed DADF approach can achieve excellent performance while maintaining good efficiency.

\subsubsection{Impact of the Dual-branch reconstruction}

\begin{figure}[t]
\centerline{\includegraphics[width=8.8cm]{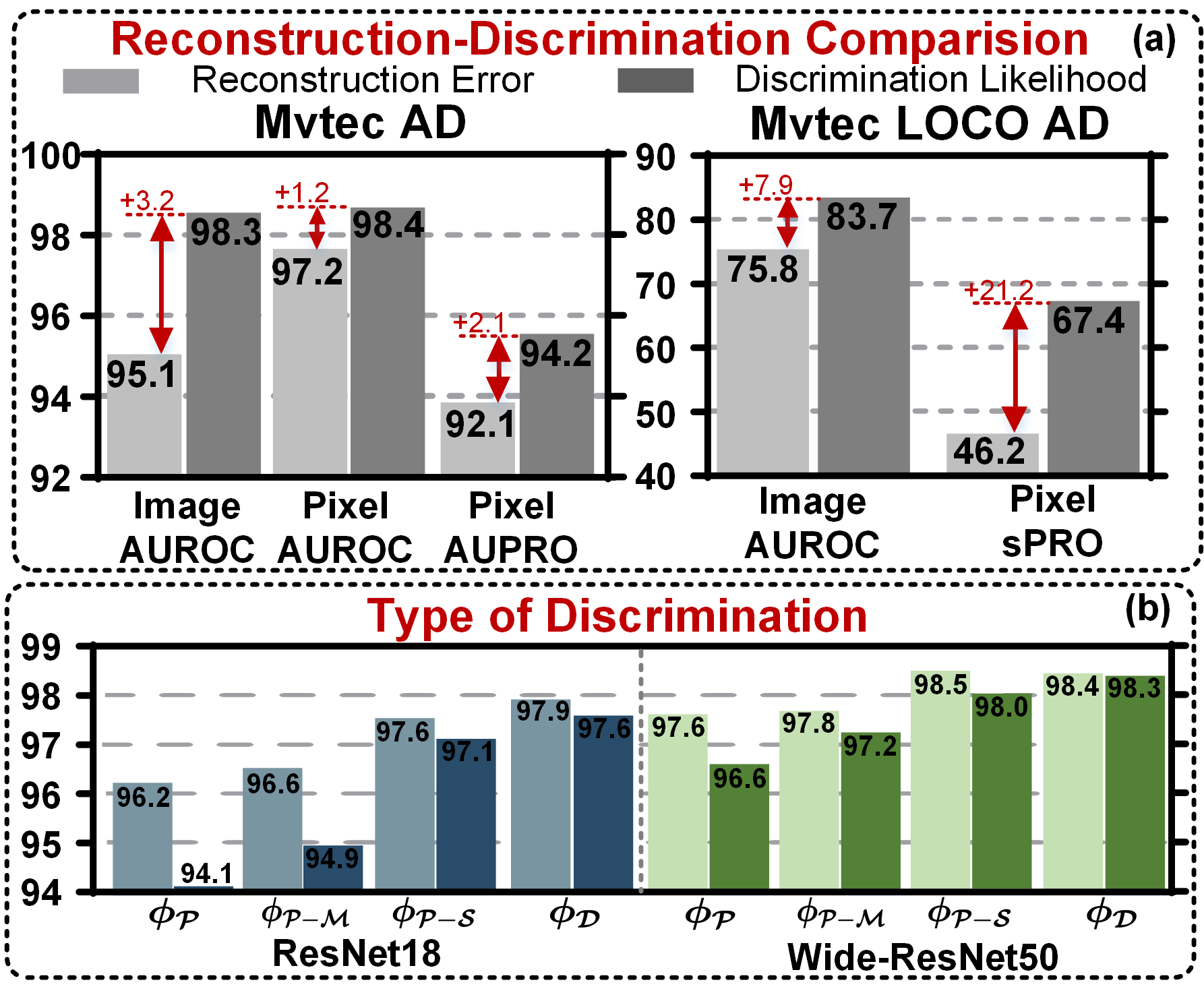}}
\caption[width=8.8cm]{
(a) The comparative analysis of the effectiveness of methods relying on reconstruction error and discriminantion likelihood on the Mvtec AD and Mvtec LOCO AD datasets. (b)Performance comparison of different types of discriminative flow.
}
\label{fig1}
\end{figure}

 One of the novel contributions presented in this paper is the dual-attention Transformer model, which uses self-attention and memorial-attention pathways to achieve stable two-level reconstruction. We demonstrate the benefits of this structure in Fig. 6(a), where we show several anomaly samples, their extracted prior features, and the two two-level reconstruction results. Our experimental results show that the reconstruction results based on self-attention can reconstruct abnormalities well, while the proposed memorial attention-based reconstruction can repair abnormal parts and produce reconstruction results that conform to normality prior. Moreover, it is noteworthy that the memory reconstruction path can suppress global anomalies such as deletions and dislocations, indicating its strong ability to understand high-level global semantics.

As stated earlier, the reconstruction error obtained can serve as an effective anomaly criterion. Fig. 6(b) presents the performance of this approach, highlighting the advantages of memory-based reconstruction in improving detection performance by +2.6/+1.2/+2.8 compared to traditional self-reconstruction. Further improvements are observed when the two methods are weightedly fused. This is attributed to the fact that self-reconstruction retains more detailed texture information, resulting in fewer false positives, despite its inability to repair abnormal parts. In contrast, memory-based reconstruction yields a more stable anomaly recovery ability, at the cost of losing detailed texture information, leading to fewer false negatives but more false positives. Hence, the use of dual reconstruction yields superior results.

\begin{figure}[t]
\centerline{\includegraphics[width=8.8cm]{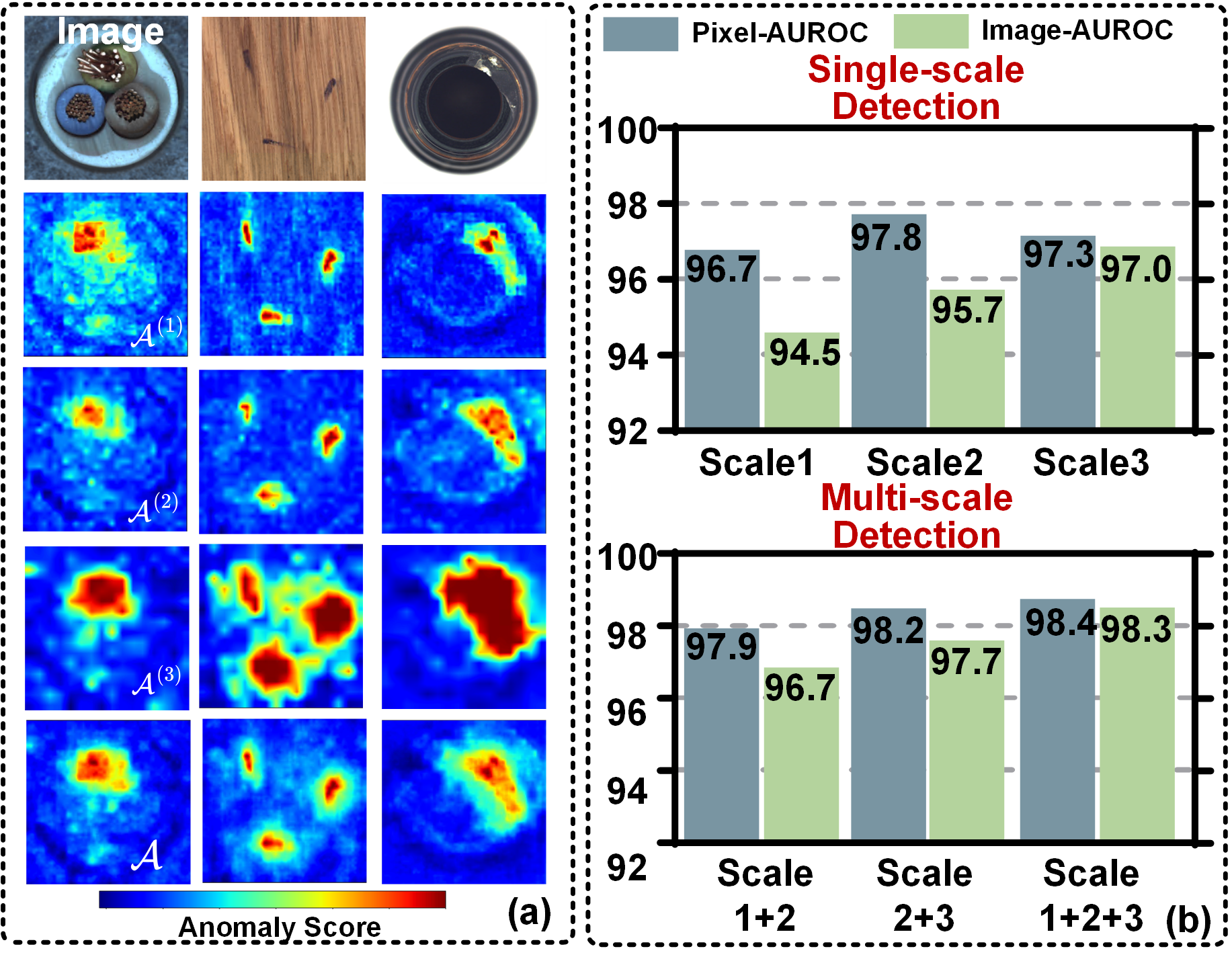}}
\caption[width=8.8cm]{
(a) Visualization results of detection effects at different scales. (b) The quantitative detection results of Different single-scale and multi-scale.
}
\label{fig1}
\end{figure}

\subsubsection{Effect of the discriminative flow}

In this study, we also present the concept of discriminative flow as a means of discrimination normality likelihood modeling. The approach offers two key benefits. Firstly, it eliminates the direct reconstruction discrepancy and replaces it with the likelihood as a more effective method of normality estimation. Secondly, it incorporates the reconstruction discrepancy as an improved discriminant property, thus enhancing the reliability of normality likelihood estimates for NF. In Fig. 7(a), a comparison is made between the discrimination likelihood method and reconstruction error-based methods. The findings demonstrate that the utilization of likelihood modeling yields a significant performance improvement compared to the reconstruction method. Specifically, the former method yields a gain of +3.2/+1.2/+2.1 on the MvtecAD. While on the Mvtec LOCO AD, the performance gain is even more significant, with gains of +7.9 and +21.2, respectively, demonstrating a considerable advantage of the former method.

In this study, we delve deeper into the impact of discriminative flow types on performance. In Figure 7(b), we identify four methods for estimating normality, denoted as $\boldsymbol{\phi_{\mathcal{P}}}$, $\boldsymbol{\phi_{\mathcal{P-S}}}$, $\boldsymbol{\phi_{\mathcal{P-M}}}$, and $\boldsymbol{\phi_{\mathcal{D}}}$, respectively representing the vanilla approach that solely utilizes prior features, the method that integrates prior and self-reconstruction results, the method that incorporates prior and memory reconstruction results, and the proposed joint method of prior and two-level reconstructions. Our experimental results demonstrate that the methods using reconstruction discrepancy discrimination can achieve varying degrees of performance improvement compared to the vanilla $\boldsymbol{\phi_{\mathcal{P}}}$ flow model. This improvement is particularly pronounced when utilizing the lightweight backbone network ResNet18, as the network's insufficient ability to differentiate between normal and abnormal instances can be supplemented by the reconstruction discrepancies, a crucial criterion for determining normality. Specifically, the method employing $\boldsymbol{\phi_{\mathcal{D}}}$ yields a performance gain of +1.7/+3.5 and +0.8/+1.8 compared to the baseline $\varphi_{\mathcal{P}}$ for the image/pixel AUROC metrics.


\subsubsection{The gain of multiscale detection}

Prior studies\cite{r11} have demonstrated that detection accuracy can be enhanced by utilizing multi-scale hierarchical techniques. Consistent with this principle, the present study extracts prior features at various scales and constructs discriminative models for each. In this section, we evaluate the detection performance of different scales.

Visualizations presented in Fig. 8(a) indicate that deep features produce more discriminative anomaly score maps, but with imprecise localization due to their lower resolution. Conversely, shallow features yield more precise anomaly localization maps, albeit with weaker anomaly/normal discriminative capabilities.

Quantitative comparisons depicted in Fig. 8(b) reveal that scale 3 achieves the best detection result, outperforming scale 1 and scale 2. Moreover, multi-scale fusion results in better performance than single-scale. The overall best performance is achieved through the fusion of all scale features.

\section{Conclusion}
This article presents the DADF, an innovative framework that combines correspondence and discriminative mechanisms. The framework's main contribution is its multi-scale pre-trained prior feature embedding, which is followed by a stable two-level reconstruction using a dual-attention Transformer. Moreover, discriminative flows are designed for each scale, and the method is extensively validated through experiments, demonstrating exceptional performance. Nonetheless, we have observed that utilizing the concatenation of prior and reconstructed features as the input of the discriminative flow directly can have an impact on computational efficiency. We consider this an area for potential future enhancement.

\bibliographystyle{ieeetr} 
\bibliography{reference}
\end{document}